\documentclass[letterpaper, 10 pt, conference]{ieeeconf}
\IEEEoverridecommandlockouts
\usepackage{graphicx}
\usepackage{amsmath,amssymb}
\usepackage{booktabs}
\usepackage{cite}
\usepackage{url}
\usepackage[hidelinks]{hyperref}
\usepackage{tikz}
\usetikzlibrary{arrows.meta,positioning,calc,fit,backgrounds}

\title{\LARGE \bf
Identifying Habit, Physics, and Nuisance\\
in Robot World Models
}

\author{Jinting Hang$^{\ast}$ and Zhenhui Cai%
\thanks{$^{\ast}$Corresponding author:
\texttt{jinting.hang@harvest-praxis.com}.
Z.~Cai: \texttt{philip.cai@harvest-praxis.com}.}%
\\
\normalsize Harvest Praxis%
}

\begin{document}
\maketitle
\thispagestyle{empty}
\pagestyle{empty}

\begin{abstract}
Teleoperated demonstrations are often multimodal even when the underlying dynamics are nearly deterministic given the executed action.
We argue that this multimodality typically mixes three factors---operator \emph{habit} in action selection, shared \emph{physics}, and observation \emph{nuisance}---and that entangled next-observation predictors absorb all three.
We formalize the split with a structural causal model $a{=}g(h,z,u)$, $z'{=}f(z,a)$, $o{=}r(z,c)$, and test it with complementary interventions: replacing or shuffling actions at fixed state sharply increases next-state error, whereas appearance and camera changes should not; habit-aware reverse scoring improves ranking of feasible pasts without rewriting the dynamics.
The associated adaptation rule is to freeze a shared physics readout and update only a thin interface.
On StackCube, DROID, and RH20T this rule improves low-shot transfer relative to training from scratch, retains cleaner dynamics under corrupted adaptation data, and extends from proprioception to pixel observations with multi-view and multi-step checks.
We do not equate latent actions with operator habit, and we do not target large-scale video generation benchmarks.
\end{abstract}

\section{INTRODUCTION}

World models and imitation learners trained on human teleoperation frequently face a familiar ambiguity: predictive distributions over the next state or image look multimodal, yet much of that diversity may come from how different operators choose actions, not from stochastic physics.
A second ambiguity is observational.
Changes in lighting, background, or camera pose alter pixels without changing the robot's configuration or the mechanical consequences of an action.
When a single predictor is fit to $P(o'|o)$~\cite{hafner2020dreamer,ha2018worldmodels}, habit, physics, and nuisance are easily entangled.
The practical symptoms are brittle adaptation and finetuning that rewrites dynamics to match a new user or camera.

\begin{figure}[t]
  \centering
\begin{tikzpicture}[
  font=\scriptsize,
  >=Stealth,
  every node/.style={align=center},
  tinybox/.style={
    draw=black!55, line width=0.4pt, rounded corners=0.5pt,
    fill=white, inner sep=0.55pt, font=\tiny,
    minimum width=0.30cm, minimum height=0.24cm
  },
  arr/.style={->, line width=0.5pt, draw=black!70},
  title/.style={font=\tiny\bfseries},
  foot/.style={font=\tiny, text=black!60, align=center}
]
  \draw[draw=black!70, line width=0.65pt, rounded corners=1.5pt, fill=black!3]
    (0.00,-2.20) rectangle (2.00,0);
  \draw[draw=black!70, line width=0.65pt, rounded corners=1.5pt, fill=cyan!6]
    (2.35,-2.20) rectangle (5.45,0);
  \draw[draw=black!70, line width=0.65pt, rounded corners=1.5pt, fill=green!6]
    (5.80,-2.20) rectangle (7.90,0);

  \draw[arr, black!55, line width=0.9pt] (2.00,-1.05) -- (2.35,-1.05);
  \draw[arr, black!55, line width=0.9pt] (5.45,-1.05) -- (5.80,-1.05);

  \node[title] at (1.00,-0.20) {A.~Entangled $P(o'|o)$};
  \fill[orange!40] (0.66,-0.52) circle (0.09);
  \fill[cyan!40]  (1.00,-0.52) circle (0.09);
  \fill[green!35] (1.34,-0.52) circle (0.09);
  \node[font=\tiny, text=black!70] at (1.00,-0.78) {habit + physics + $c$};
  \node[foot] at (1.00,-1.80)
    {one predictor absorbs all three\\[-0.05em]$\Rightarrow$ brittle adapt};

  \begin{scope}
    \clip (2.40,-2.15) rectangle (5.40,-0.05);
    \node[title] at (3.90,-0.20) {B.~SCM split};
    \node[tinybox, fill=green!12]  (Nc)  at (3.45,-0.52) {$c$};
    \node[tinybox, fill=green!12]  (No)  at (3.95,-0.52) {$o$};
    \node[tinybox, fill=orange!12] (Nh)  at (2.80,-0.90) {$h$};
    \node[tinybox, fill=orange!12] (Nu)  at (2.80,-1.28) {$u$};
    \node[tinybox, fill=orange!12] (Na)  at (3.30,-1.08) {$a$};
    \node[tinybox, fill=cyan!12]  (Nz)  at (3.80,-1.08) {$z$};
    \node[tinybox, fill=cyan!12]  (Nzp) at (4.30,-1.08) {$z'$};
    \draw[arr, orange!75!black] (Nh) -- (Na);
    \draw[arr, orange!75!black] (Nu) -- (Na);
    \draw[arr, cyan!70!black] (Na) to[bend right=8] (Nzp);
    \draw[arr, cyan!70!black] (Nz) -- (Nzp);
    \draw[arr, green!55!black] (Nc) -- (No);
    \draw[arr, green!55!black] (Nz) -- (No);
    \draw[->, dashed, red!60!black, line width=0.45pt]
      (Nh) to[bend left=12]
      node[midway, above, font=\tiny\itshape, text=red!60!black, inner sep=0.08pt] {no}
      (Nz);
    \node[foot] at (3.90,-1.95) {intervene on $a$ vs.\ $c$};
  \end{scope}

  \node[title] at (6.85,-0.20) {C.~Freeze + interface};
  \node[draw=cyan!60!black, fill=cyan!10, rounded corners=1pt,
        minimum width=1.55cm, minimum height=0.34cm, font=\tiny]
        (Cf) at (6.85,-0.65) {freeze $f^\star$ / Enc};
  \node[draw=orange!70!black, fill=orange!12, rounded corners=1pt,
        minimum width=1.55cm, minimum height=0.34cm, font=\tiny]
        (Cg) at (6.85,-1.15) {adapt $g$ / $\delta$};
  \draw[arr, black!65] (Cf) -- (Cg);
  \node[foot] at (6.85,-1.85)
    {low-$N$: freeze $\preceq$ scratch,\\[-0.05em]$\approx$ finetune};
\end{tikzpicture}
  \caption{From entangled next-observation predictors (A) to an SCM that separates habit, physics, and nuisance (B), and a freeze-and-adapt interface (C).}
  \label{fig:claim}
\end{figure}
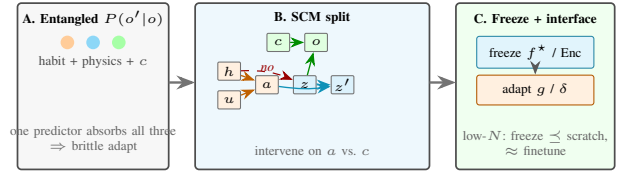

This paper treats the ambiguity as an identification problem.
We posit a structural causal model (SCM) in which habit influences the future only through actions, nuisance influences only the rendering of observations, and physics is a shared map $z'=f(z,a)$ (Fig.~\ref{fig:claim}).
Under that graph, action-level interventions should stress the physics readout, while appearance and viewpoint interventions should not.
If the graph is approximately right and the state is sufficiently complete, few-shot adaptation should update a thin policy or residual interface while freezing the shared law.

Our contributions are threefold.
First, we state a single SCM that separates habit, physics, and nuisance, and we organize interventions into physics-facing and habit-facing readouts, together with nuisance checks on images.
Second, we characterize when freezing dynamics is justified and when it is not, including coarse command interfaces, over-aggressive training objectives, cross-task habit transfer, and host stacks that ignore an injected dynamics factor.
Third, we evaluate a controlled freeze-versus-finetune-versus-scratch protocol on a Physical Kernel in simulation, on DROID and RH20T proprioception, and on DROID pixels with multi-view, multi-step, reverse-scoring, and low-shot adaptation experiments.

\section{STRUCTURAL CAUSAL MODEL}

We write
\begin{align}
  a_t &= g(h_t, z_t, u_t), \label{eq:g}\\
  z_{t+1} &= f(z_t, a_t), \label{eq:f}\\
  o_t &= r(z_t, c_t), \label{eq:r}
\end{align}
where $h_t$ denotes operator habit or preference, $u_t$ is a possibly lossy task or command interface, and $c_t$ collects observation nuisances such as illumination and camera identity.
Equations~\eqref{eq:g}--\eqref{eq:r} encode two exclusion restrictions: there is no direct edge from habit into the next state, and no direct edge from nuisance into the dynamics.
Images are treated as observations of $z$, not as a substitute for $f(z,a)$.
Throughout, ``dynamics'' means a discrete one-step or finite-horizon map rather than a dense video generator.

\begin{figure}[t]
  \centering
  \resizebox{0.92\linewidth}{!}{
\begin{tikzpicture}[
  font=\footnotesize,
  >=Stealth,
  node distance=1.05cm and 1.45cm,
  every node/.style={align=center},
  var/.style={
    draw=black!70,
    line width=0.7pt,
    rounded corners=1.2pt,
    minimum width=1.6cm,
    minimum height=0.72cm,
    inner sep=2pt,
    fill=#1
  },
  lab/.style={font=\scriptsize, text=#1},
  arr/.style={->, line width=0.9pt, draw=#1},
]
  \node[var=orange!12] (h) {habit $h$};
  \node[var=orange!12, below=1.1cm of h] (u) {command $u$};
  \node[var=orange!12, right=1.6cm of h, yshift=-0.55cm] (a) {action $a$};
  \node[var=cyan!12, right=1.7cm of a] (z) {state $z$};
  \node[var=cyan!12, right=1.7cm of z] (zp) {$z'$};
  \node[var=green!12, above=1.0cm of z] (c) {nuisance $c$};
  \node[var=green!12, above=1.0cm of zp] (o) {obs $o$};

  \draw[arr=orange!80!black] (h) -- (a);
  \draw[arr=orange!80!black] (u) -- (a);

  \draw[arr=cyan!70!black] (z) -- node[above, font=\scriptsize, text=cyan!60!black, inner sep=1pt] {$f$} (zp);
  \draw[arr=cyan!70!black] (a.east) to[out=-15, in=-160] (zp.south west);

  \draw[arr=green!60!black] (c) -- (o);
  \draw[arr=green!60!black] (z) -- (o);

  \draw[->, dashed, line width=0.75pt, draw=red!65!black]
    (h.east) to[out=28, in=150]
    node[midway, above, font=\scriptsize\itshape, text=red!65!black] {no}
    (z.north west);

  \node[lab=orange!80!black, above left=0.02cm and -0.05cm of h] {policy / interface};
  \node[lab=green!50!black, above=0.1cm of o] {rendering};
  \node[lab=cyan!60!black, below=0.55cm of z]
    {physics: $z'{=}f(z,a)$\;(no $h{\to}z$, no $c{\to}f$)};
\end{tikzpicture}}
  \caption{Graphical summary of~\eqref{eq:g}--\eqref{eq:r}.
  Solid arrows are allowed mechanisms; the dashed edge marks the exclusion $h\nrightarrow z$.}
  \label{fig:scm}
\end{figure}
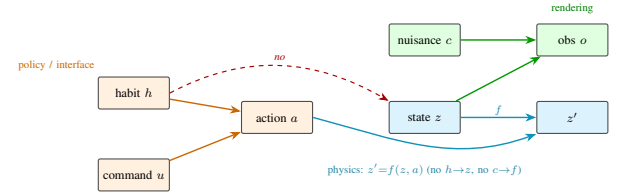

\paragraph{Interventions.}
Shuffling or replacing the action at fixed $z$ is an action-level proxy for changing habit.
Swapping operators on the same task provides a stronger within-task test: the state remains on-task, but the action is drawn from another user's policy.
Near-zero actions probe autonomous drift and should remain concentrated when $z$ is informative.
Appearance jitter and a held-out second camera proxy $do(c)$ and, under a nuisance-stable encoder, should leave structural next-joint error nearly unchanged.
Table~\ref{tab:intervene} summarizes the intended contrasts.

\begin{table}[t]
\centering
\caption{Intervention design.
Physics-facing readouts emphasize next-state or next-joint error; habit-facing readouts emphasize policy residuals and ranking.}
\label{tab:intervene}
\scriptsize
\begin{tabular}{@{}p{0.20\linewidth}p{0.20\linewidth}p{0.28\linewidth}p{0.22\linewidth}@{}}
\toprule
Intervention & Proxy & Physics readout & Habit / note \\
\midrule
Wrong action & shuffle at fixed $z$ & Kernel ${\sim}10\times$; DROID ${\sim}13\times$; RH20T ${\sim}10^{3}\times$ & action changed \\
Swap user & other operator's $a$ & RH20T ${\sim}1362\times$ & within-task habit \\
Habit fingerprint & $a{-}\hat\pi$; scramble & --- & ID $0.52{\to}0.21$ \\
Corrupted adapt.\ & $a{\times}2$ demos & freeze ${\sim}150\times$ cleaner & finetune absorbs habit \\
Appearance $do(c)$ & color / blur & pixel ratio ${\sim}0.99$ & next-RGB sensitive \\
Second view & held-out camera & cross ${\sim}1.42\times$ & harder nuisance \\
Reverse scoring & phys.$\times$habit & GT remains feasible & mass lift \\
Freeze interface & train $g$ / $\mathrm{delta}$ & low-$N$ $\le$ scratch & adaptation rule \\
\bottomrule
\end{tabular}
\end{table}

\section{METHOD}

\subsection{Proprioceptive and Kernel Models}
On ManiSkill StackCube~\cite{maniskill} we use a Physical Kernel with encoder outputs $z_p$, dynamics $f$, and a thin map $g$ from commands $u$ to actions.
On DROID and RH20T proprioception we use an MLP $f(j,a)\mapsto j'$.
Given a pretrained $f^*$, we compare three adaptation regimes at matched shot budgets $N$: freeze $f^*$ and train only $g$; finetune $f$ and $g$ jointly; and train both from scratch.

\subsection{Pixel Model}
For DROID exterior images we predict proprioceptive joints through a structural readout
\begin{equation}
  z=\mathrm{Enc}(o),\quad
  \hat j=\mathrm{probe}(z),\quad
  \hat j'=\hat j+\mathrm{delta}(z,a).
\end{equation}
The training objective aligns $\mathrm{probe}(z)$ to the current joints, regresses next joints through $\mathrm{delta}$, and applies a hinge that requires shuffled actions to increase error.
A light appearance-consistency term may be added on Enc.
We retain transitions in the upper quantile of $\|\Delta j\|$ so that idle frames do not dominate.

A failure mode is worth stating explicitly.
If one optimizes only latent consistency of the form $f(\mathrm{Enc}(o),a)\approx\mathrm{Enc}(o')$ without a joint-facing target, the representation can become nearly action-insensitive: appearance interventions look harmless while wrong actions barely hurt.
The joint-primary objective above is used to avoid that collapse.

\subsection{Multi-Step Prediction}
For horizon $H$ we unroll
\[
\hat j \leftarrow \hat j + \mathrm{delta}(\mathrm{Enc}(o_t), a_k),\qquad k=1,\ldots,H,
\]
and compare true action sequences against shuffled sequences and against appearance-perturbed observations.
Action errors should accumulate with $H$; appearance perturbations should not.

\subsection{Reverse Scoring}
Given a candidate past $(z,a)$ and an observed successor $z'$, we score
\begin{align*}
\log L_{\mathrm{phys}} &\propto -\|f(z,a)-z'\|^2,\\
\log L_{\mathrm{hab}} &\propto -\|a-g(h,z,u)\|^2,
\end{align*}
and compare physics-only scores with scores that include a correct or incorrect habit prior.
A useful habit prior should raise posterior mass on the ground-truth past and favor the correct user's $g$ over an incorrect user's, without moving the ground truth outside the physics-feasible set.

\subsection{Scope}
Latent-action methods that recover $\tilde a$ from video~\cite{schmidt2024lapo,adaworld} address a different object than operator habit in~\eqref{eq:g}.
We likewise do not study theory-of-mind world models or large video-generation leaderboards.
External video--action adapters appear only as a boundary case in Sec.~\ref{sec:identify}.

\section{EXPERIMENTAL SETUP}

\paragraph{Data.}
RH20T supplies multi-operator proprioceptive transitions~\cite{fang2024rh20t}.
DROID supplies in-the-wild manipulation episodes with exterior RGB, 7-DoF actions, and joint states~\cite{khazatsky2024droid}; we use one exterior stream for training and a second exterior stream for cross-camera evaluation.
StackCube experiments use the Physical Kernel protocol in ManiSkill~\cite{maniskill}.

\paragraph{Protocol.}
Pixel, multi-step, downstream, and reverse-scoring experiments on DROID are reported as mean$\pm$std over seeds $\{0,1,2\}$ under fixed episode and epoch budgets.
Proprioceptive and Kernel results follow the same freeze / finetune / scratch template.

\paragraph{Metrics.}
We report next-state or next-joint mean squared error ratios under action and appearance interventions, operator identification from phase-level action residuals, reverse-scoring mass and pairwise win rates, low-shot adaptation error, and StackCube stacked success for closed-loop checks.

\section{HABIT VERSUS PHYSICS}

\begin{figure}[t]
  \centering
  \includegraphics[width=0.88\linewidth]{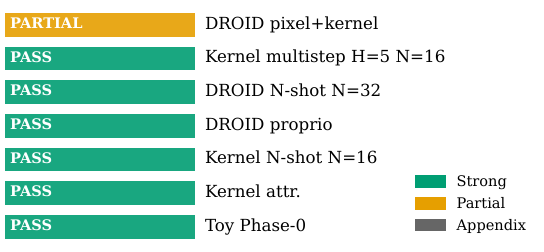}
  \caption{Compact status summary of the main diagnosis and transfer checks.}
  \label{fig:ladder}
\end{figure}

\begin{table}[t]
\centering
\caption{Selected quantitative results.
``Partial'' denotes a weak but non-null effect; ``weak'' denotes an inconclusive cross-task test.}
\label{tab:ladder}
\scriptsize
\begin{tabular}{@{}p{0.42\linewidth}p{0.35\linewidth}c@{}}
\toprule
Setting & Metric & Outcome \\
\midrule
Kernel $z_p$ mix / true & ${\sim}10.4\times$ & strong \\
DROID proprio wrong / true & ${\sim}13\times$ & strong \\
RH20T swap / shuffle & ${\sim}1362\times$ / ${\sim}1352\times$ & strong \\
RH20T LOUO $f_{\mathrm{all}}/f_{\mathrm{self}}$ & $0.38\times$ & strong \\
Fingerprint ID / scramble & $0.52{\to}0.21$ ($-31$pp) & strong \\
Corrupted adapt., freeze vs.\ finetune & ${\sim}150\times$ cleaner & strong \\
Low-shot freeze vs.\ scratch & freeze better at low $N$ & strong \\
StackCube closed loop @ $N{=}128$ & $0.22$ vs.\ $0.10$ & strong \\
Cross-task habit / matched state & ${\approx}1.0\times$ / $S{\gtrsim}H$ & weak \\
Frozen sim encoder on DROID pixels & mix/true ${\approx}1.09$ & partial \\
\bottomrule
\end{tabular}
\end{table}

\subsection{StackCube Kernel}
\begin{figure}[t]
  \centering
  \begin{minipage}[t]{0.48\linewidth}
    \centering
    \includegraphics[width=\linewidth]{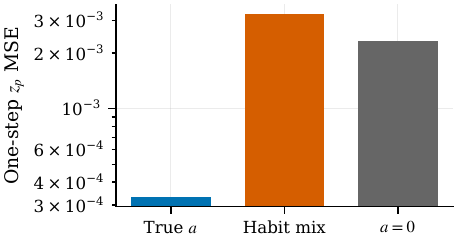}
    \centerline{\scriptsize (a) Attribution}
  \end{minipage}\hfill
  \begin{minipage}[t]{0.48\linewidth}
    \centering
    \includegraphics[width=\linewidth]{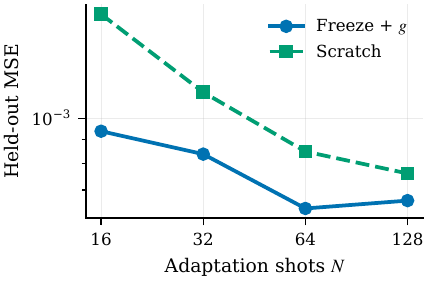}
    \centerline{\scriptsize (b) Low-shot adaptation}
  \end{minipage}
  \caption{StackCube Kernel: (a) wrong-action / idle attribution; (b) freeze+$g$ vs.\ finetune vs.\ scratch.}
  \label{fig:kerattr}
\end{figure}

Replacing the true action by a mixture of alternatives increases one-step $z_p$ error by about $10.4\times$ (Fig.~\ref{fig:kerattr}a).
At fixed initial state, habit-induced action mixtures produce action-conditioned branching, whereas holding $a{=}0$ remains concentrated.
With $N{=}16$ adaptation transitions, freezing $f^*$ and training $g$ yields error $0.00089$, below joint finetuning ($0.00115$) and scratch ($0.00169$) (Fig.~\ref{fig:kerattr}b).
The same ordering holds for horizon-$5$ rollouts.

As a closed-loop sanity check we adapt freeze+$g$ on StackCube with $N{=}128$ demonstrations and evaluate stacked success over $40$ episodes.
A privileged full-policy ceiling reaches $0.80$ with latent roll and $0.65$ with per-step re-encoding.
Freeze+$g$ attains $0.22{\pm}0.09$ versus $0.10{\pm}0.07$ for scratch across eight seeds (better on $7/8$).
When adaptation demos are deliberately scaled by $a{\times}2$, freeze+$g$ remains ahead of scratch ($0.24{\pm}0.09$ versus $0.13$ over four seeds).

\subsection{DROID Proprioception}
\begin{figure}[t]
  \centering
  \begin{minipage}[t]{0.48\linewidth}
    \centering
    \includegraphics[width=\linewidth]{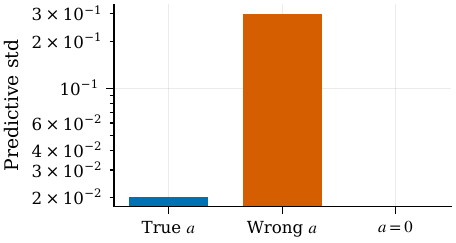}
    \centerline{\scriptsize (a) Attribution}
  \end{minipage}\hfill
  \begin{minipage}[t]{0.48\linewidth}
    \centering
    \includegraphics[width=\linewidth]{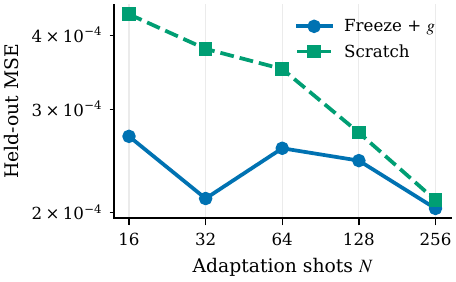}
    \centerline{\scriptsize (b) In-domain $f_{\mathrm{robot}}$}
  \end{minipage}
  \caption{DROID proprioception: (a) predictive spread; (b) freeze+$g$ on in-domain dynamics.}
  \label{fig:droidattr}
\end{figure}

On DROID joints, wrong-action predictive spread is about $0.28$ against $0.02$ for the true action, while idle spread is near $6{\times}10^{-4}$ (Fig.~\ref{fig:droidattr}a).
At $N{=}32$, freeze+$g$ matches or beats finetuning and improves on scratch ($0.00030$ versus $0.00039$).
Retraining an in-domain residual model $f_{\mathrm{robot}}$ with a moderate wrong-action hinge, and without copying StackCube weights, restores large motion sensitivity and low-shot freeze gains for $N\in\{16,\ldots,256\}$ (Fig.~\ref{fig:droidattr}b).
This supports reuse of the \emph{protocol}, not transfer of a frozen simulator dynamics network: Paths that transplant StackCube $f$ into DROID remain only weakly action-sensitive (mix/true about $1$--$1.3$).

\subsection{RH20T Multi-Operator Data}
A leave-one-user pooled dynamics model approaches same-user next-state error ($f_{\mathrm{all}}/f_{\mathrm{self}}\approx0.38$), whereas fitting on a single other user is unstable (about $3.9\times$).

On a shared $f$, pairing the true state with another operator's action increases MSE by roughly $1362\times$, essentially matching unstructured action shuffle ($1352\times$; Table~\ref{tab:rh20t}).
Thus habit can be manipulated without leaving the task.
Phase-level residuals $a-\hat\pi(s)$ identify operators at accuracy $0.52$ (majority baseline about $0.25$); scrambling the residual fingerprint drops accuracy to $0.21$.
Concatenating raw state into the classifier is inconclusive, because state trajectories already carry stylistic information.
We therefore do not require physics residuals alone to be identity-free: under~\eqref{eq:f}, residuals may legitimately reflect $a$.

\begin{table}[t]
\centering
\caption{RH20T interventions on the action argument of a shared dynamics model.}
\label{tab:rh20t}
\small
\begin{tabular}{@{}lrrr@{}}
\toprule
Condition & MSE & vs.\ true & Role \\
\midrule
true $a$ & $2.09{\times}10^{-7}$ & $1\times$ & reference \\
swap-user $a$ & $2.84{\times}10^{-4}$ & ${\sim}1362\times$ & within-task habit \\
shuffle $a$ & $2.82{\times}10^{-4}$ & ${\sim}1352\times$ & unstructured action \\
$a{=}0$ & $1.41{\times}10^{-4}$ & --- & idle \\
\bottomrule
\end{tabular}
\end{table}

\begin{figure}[t]
  \centering
  \begin{minipage}[t]{0.48\linewidth}
    \centering
    \includegraphics[width=\linewidth]{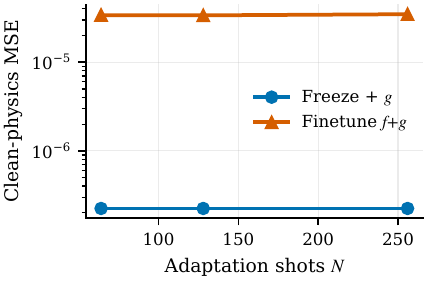}
    \centerline{\scriptsize (a) Corrupted adaptation}
  \end{minipage}\hfill
  \begin{minipage}[t]{0.48\linewidth}
    \centering
    \includegraphics[width=\linewidth]{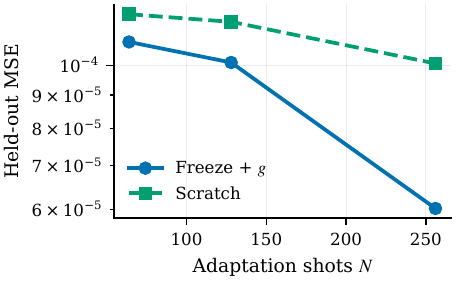}
    \centerline{\scriptsize (b) Mid-lossy $u$}
  \end{minipage}
  \caption{RH20T: (a) freeze under $a{\times}2$ demos; (b) freeze+$g$ with mid-lossy $u$.}
  \label{fig:expe}
\end{figure}

When adaptation uses over-forced demonstrations ($a{\times}2$), freezing $f$ and training $g$ preserves clean-physics next-state error by about two orders of magnitude relative to finetuning $f$ (Fig.~\ref{fig:expe}a): finetuning absorbs the corrupted habit into the law.
With a moderately informative command interface $u$, freeze+$g$ beats scratch for $N\ge64$ on nearly all splits (Fig.~\ref{fig:expe}b).
An excessively coarse $u$ can make the same protocol look unsuccessful even when oracle-action dynamics remain accurate; that is a false negative for the interface, not evidence against shared physics.

\subsection{Synthetic Flow Matching and Training Objectives}
\begin{figure}[t]
  \centering
  \includegraphics[width=0.80\linewidth]{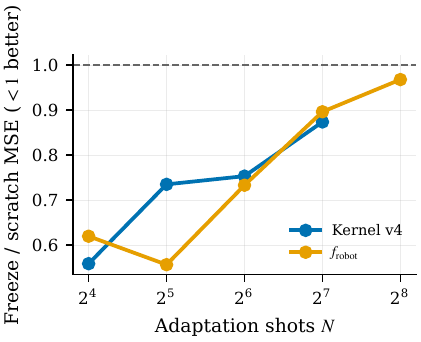}
  \caption{Effect of dynamics training choices on low-shot freeze+$g$ transfer.}
  \label{fig:nshot}
\end{figure}

A conditional OT-CFM toy recovers the same qualitative picture: shuffling actions inflates loss by about $60\times$, and freezing the physics velocity field while adapting a thin $g$ helps for $N\in\{5,\ldots,100\}$.
Figure~\ref{fig:nshot} compares dynamics trained with different wrong-action hinges.
A mid-strength hinge keeps both large mix/true ratios and low-shot freeze gains.
An over-hard hinge can preserve a large diagnostic ratio while destroying low-shot transfer.
An in-domain robot model recovers both.
Action sensitivity is therefore necessary but not sufficient for a useful few-shot base.

\section{NUISANCE, VISION, AND REVERSE SCORING}

\subsection{Appearance and Second-Camera Checks}
On DROID RGB with the joint-primary model, appearance interventions change next-joint error by a factor $0.988{\pm}0.054$, while wrong actions increase it by $9.5{\pm}3.7\times$ (seeds $\{0,1,2\}$).
An entangled next-image baseline remains sensitive to the same appearance edits.
Training on one exterior camera and evaluating on a second yields appearance factor $0.984{\pm}0.067$ and cross-camera factor $1.42{\pm}0.06$ (Table~\ref{tab:vision}).
Cross-camera shift is harder than photometric jitter, yet still far smaller than action interventions.
Borrowing a frozen simulator encoder without joint-primary training remains only weakly action-sensitive (about $1.09$), which is consistent with needing an in-domain structural readout rather than an off-the-shelf visual backbone alone.

\begin{table}[t]
\centering
\caption{DROID vision and nuisance summary (mean$\pm$std over three seeds).}
\label{tab:vision}
\small
\begin{tabular}{@{}lc@{}}
\toprule
Quantity & Value \\
\midrule
Appearance $do(c)$ ratio & $0.988{\pm}0.054$ \\
Second-camera ratio & $1.42{\pm}0.06$ \\
Wrong-action ratio (structural) & $9.5{\pm}3.7\times$ \\
Multi-step wrong-action ($H{=}1/5/10$) & $1.9$ / $7.3$ / $13.7$ \\
Downstream freeze MSE at $N{=}32$ & $0.194{\pm}0.018$ \\
Reverse mass lift vs.\ physics-only & $+0.040{\pm}0.002$ \\
\bottomrule
\end{tabular}
\end{table}

\subsection{Multi-Step Prediction and Reverse Scoring}
\begin{figure}[t]
  \centering
  \begin{minipage}[t]{0.48\linewidth}
    \centering
    \includegraphics[width=\linewidth]{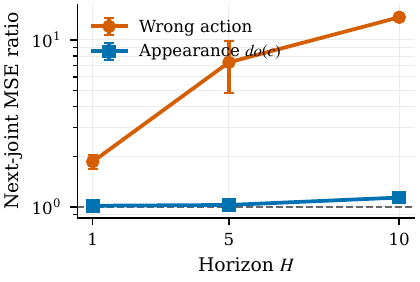}
    \centerline{\scriptsize (a) Multi-step ratios}
  \end{minipage}\hfill
  \begin{minipage}[t]{0.48\linewidth}
    \centering
    \includegraphics[width=\linewidth]{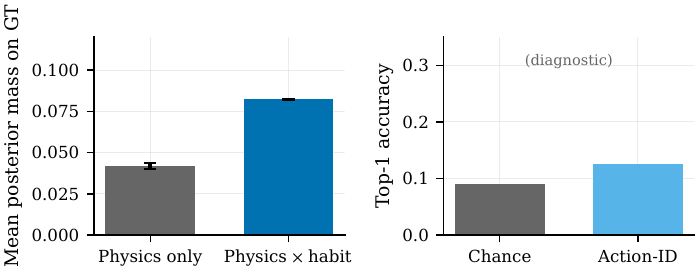}
    \centerline{\scriptsize (b) Reverse scoring}
  \end{minipage}
  \caption{(a) Wrong-action ratios grow with $H$ while appearance $do(c)$ stays near one.
  (b) Habit raises GT posterior mass; action-ID remains weak.}
  \label{fig:hcurve}
\end{figure}
For horizons $H\in\{1,5,10\}$, wrong-action ratios are $1.9{\pm}0.2$, $7.3{\pm}2.5$, and $13.7{\pm}0.6$, while appearance ratios stay near one (Fig.~\ref{fig:hcurve}a).
Longer horizons amplify action mistakes without converting photometric nuisance into apparent dynamics.
Including the correct habit prior increases mean posterior mass on the ground-truth past by $0.040{\pm}0.002$ relative to physics-only scoring, and the correct user's $g$ beats a wrong user's on $0.649{\pm}0.015$ of trials (Fig.~\ref{fig:hcurve}b).
Direct user identification from action style remains close to chance (about $13\%$ top-1 versus $9\%$).
We therefore treat reverse scoring as a ranking aid, not as hard operator recognition.

\subsection{Pixel Low-Shot Adaptation}
\begin{figure}[t]
  \centering
  \includegraphics[width=0.78\linewidth]{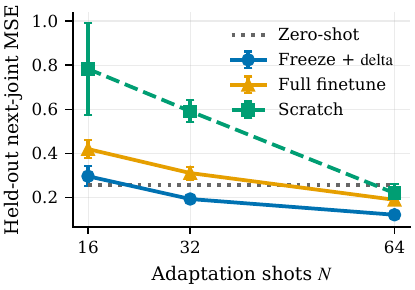}
  \caption{Pixel low-shot next-joint error after freezing Enc and probe (mean$\pm$std over seeds).}
  \label{fig:down}
\end{figure}
\begin{table}[t]
\centering
\caption{Pixel low-shot next-joint MSE (seed~0).
Across three seeds, freeze beats scratch on a majority of budgets; at $N{=}32$ freeze error is $0.194{\pm}0.018$.}
\label{tab:down}
\small
\begin{tabular}{cccc}
\toprule
$N$ & freeze+$\mathrm{delta}$ & finetune & scratch \\
\midrule
16 & 0.29 & 0.38 & 1.01 \\
32 & 0.22 & 0.35 & 0.65 \\
64 & 0.11 & 0.19 & 0.18 \\
\bottomrule
\end{tabular}
\end{table}
Freezing Enc and probe while adapting only $\mathrm{delta}$ improves over scratch at low $N$ and remains competitive with full finetuning (Fig.~\ref{fig:down}, Table~\ref{tab:down}).
A proprioceptive oracle that never uses pixels reaches about $0.007$ next-joint MSE, indicating that images here act as an observation channel into the same structural target rather than as a replacement for joint physics.

\section{WHEN FREEZING DYNAMICS IS JUSTIFIED}
\label{sec:identify}

Negative and weak outcomes are part of the identification story.

A usable command interface matters.
On RH20T, an overly coarse $u$ can make freeze+$g$ look ineffective even though dynamics conditioned on oracle actions remain accurate.
That failure diagnoses the interface, not the absence of shared physics.

Diagnostic sensitivity is not the same as transfer quality.
Objectives that aggressively maximize wrong-action ratios can harm low-shot freeze+$g$, whereas moderate hinges and in-domain robot models preserve both (Fig.~\ref{fig:nshot}).

Representation domain matters.
Frozen simulator encoders on DROID images remain only partially action-sensitive; proprioception, in-domain dynamics, and joint-primary pixel models carry the robot evidence.

Habit, as isolated here, is within-task.
Swap-user interventions work on a shared task (Table~\ref{tab:rh20t}), but leave-one-task-out style transfer is inconclusive (win rate about $57\%$, ratio near $1$).
State-matched neighborhoods likewise leave substantial operator information in trajectories.
We therefore intervene on actions rather than demand state-conditional purity, and we do not claim a cross-task personality code.

If habit altered contact parameters directly, an $h\to z$ edge would appear and freezing $f$ would be misspecified.
Conversely, if a downstream controller never reads an injected $f$, scrambling that factor leaves control error unchanged: the host must consume dynamics for the SCM prescription to matter.
Open-loop probes on an external DreamZero-style action model illustrate a related boundary (Tables~\ref{tab:dit} and~\ref{tab:openloop}): visual and language channels shift actions, while a small LoRA adapter underperforms the base and a frozen-kernel bridge merely ties it.
Those comparisons motivate freeze-versus-finetune-versus-scratch on a shared $f^*$, not an entangled adapter as the main claim.

\begin{table}[t]
\centering
\caption{External action-model probes (boundary experiments, not the proposed method).}
\label{tab:dit}
\small
\begin{tabular}{@{}ll@{}}
\toprule
Probe & Result \\
\midrule
Seed action std & $0.79$ \\
Video shuffle / true MSE & $5.0\times$ \\
Language-induced $\overline{\Delta a^2}$ & $0.033$ \\
State shuffle / true MSE & $1.00\times$ \\
\bottomrule
\end{tabular}
\end{table}

\begin{table}[t]
\centering
\caption{Matched open-loop action MSE on the external model.}
\label{tab:openloop}
\scriptsize
\begin{tabular}{@{}llcc@{}}
\toprule
Method & $N$ & MSE$\pm$SE & $\Delta$ vs.\ base \\
\midrule
Base & --- & $0.098\pm0.018$ & --- \\
LoRA & 16 & $0.107\pm0.021$ & $+0.010$ \\
Kernel bridge & 16 & $0.096\pm0.018$ & $-0.001$ \\
\bottomrule
\end{tabular}
\end{table}

\section{RELATED WORK}

\paragraph{World models and visual dynamics.}
Latent dynamics models such as Dreamer and related world-model agents learn from imagined rollouts~\cite{hafner2020dreamer,hafner2021dreamerv2,hafner2023dreamerv3,ha2018worldmodels}.
Visual foresight couples perception with future prediction for control~\cite{finn2017vision,ebert2018visual}.
Our question is when multi-operator, multi-camera demos justify freezing a shared $f$ rather than fitting an entangled $P(o'|o)$.

\paragraph{Latent actions and inverse dynamics.}
Methods that recover $\tilde a$ from observational video---including LAPO and AdaWorld---learn controllable bottlenecks~\cite{schmidt2024lapo,adaworld,edwards2019imitating,baker2022video}.
Inverse-/forward-dynamics pairs are classical tools for representation learning~\cite{jordan1992forward,pathak2017curiosity}.
We keep executed actions as intervention handles and treat $\tilde a$ as distinct from operator habit in~\eqref{eq:g}.

\paragraph{Imitation, preferences, and adapters.}
Behavior cloning fits action distributions from demonstrations~\cite{pomerleau1989alvinn,ross2011dagger,florence2022ibc}.
Preference-conditioned and hierarchical policies address heterogeneity~\cite{christiano2017preferences,ajay2023compositional}, but seldom publish freeze-$f$ diagnostics under $do(c)$.
Parameter-efficient adapters such as LoRA are practical cousins of thin interfaces~\cite{hu2022lora}.

\paragraph{Invariance, causality, and corpora.}
Domain randomization and view-invariant learning address nuisance variation~\cite{tobin2017domain,sermanet2018time}.
Causal perspectives motivate intervention-based evaluation~\cite{scholkopf2021causal,pearl2009causality}.
We use public manipulation resources including ManiSkill, DROID, RH20T, Open X-Embodiment, and BridgeData~\cite{maniskill,khazatsky2024droid,fang2024rh20t,o2024openx,walke2023bridgedata}.

\section{LIMITATIONS}

Our interventions remain proxies for latent habit; we do not train an explicit habit identifier as a product.
Cross-camera degradation is larger than photometric nuisance, reverse scoring helps ranking more than hard identity, and joint readouts are not video generators.
Closed-loop evidence is simulated; hardware deployment is left open.
Public offline corpora support the present claims, but they do not replace on-robot validation.

\section{CONCLUSION}

Operator habit, shared physics, and observation nuisance can be separated, at least approximately, by a simple SCM and a small family of interventions.
When the exclusions hold, freezing a shared dynamics or state readout and adapting a thin interface is a more disciplined response to scarce or corrupted demos than rewriting the law.
The supporting evidence spans simulation, multi-operator proprioception, and DROID pixels, together with explicit cases where the prescription fails for identifiable reasons.

\section*{ACKNOWLEDGMENT}
The authors thank colleagues at Harvest Praxis for discussions.


\begin{thebibliography}{99}

\bibitem{hafner2020dreamer}
D.~Hafner, T.~Lillicrap, J.~Ba, and M.~Norouzi,
``Dream to control: Learning behaviors by latent imagination,''
in \emph{ICLR}, 2020.

\bibitem{hafner2021dreamerv2}
D.~Hafner, T.~P.~Lillicrap, M.~Norouzi, and J.~Ba,
``Mastering Atari with discrete world models,''
in \emph{ICLR}, 2021.

\bibitem{hafner2023dreamerv3}
D.~Hafner, J.~Pasukonis, J.~Ba, and T.~Lillicrap,
``Mastering diverse domains through world models,''
arXiv:2301.04104, 2023.

\bibitem{ha2018worldmodels}
D.~Ha and J.~Schmidhuber,
``World models,''
arXiv:1803.10122, 2018.

\bibitem{finn2017vision}
C.~Finn and S.~Levine,
``Deep visual foresight for planning robot motion,''
in \emph{ICRA}, 2017.

\bibitem{ebert2018visual}
F.~Ebert, C.~Finn, S.~Dasari, A.~Xie, A.~Lee, and S.~Levine,
``Visual foresight: Model-based deep reinforcement learning for vision-based robotic control,''
arXiv:1812.00568, 2018.

\bibitem{schmidt2024lapo}
D.~Schmidt and M.~Jiang,
``Learning to act without actions,''
in \emph{ICLR}, 2024.

\bibitem{adaworld}
S.~Gao, S.~Zhou, Y.~Du, J.~Zhang, and C.~Gan,
``AdaWorld: Learning adaptable world models with latent actions,''
in \emph{ICML}, 2025.

\bibitem{edwards2019imitating}
A.~Edwards, H.~Sahni, Y.~Schroecker, and C.~Isbell,
``Imitating latent policies from observation,''
in \emph{ICML}, 2019.

\bibitem{baker2022video}
B.~Baker \emph{et al.},
``Video PreTraining (VPT): Learning to act by watching unlabeled online videos,''
in \emph{NeurIPS}, 2022.

\bibitem{jordan1992forward}
M.~I.~Jordan and D.~E.~Rumelhart,
``Forward models: Supervised learning with a distal teacher,''
\emph{Cognitive Science}, 1992.

\bibitem{pathak2017curiosity}
D.~Pathak, P.~Agrawal, A.~A.~Efros, and T.~Darrell,
``Curiosity-driven exploration by self-supervised prediction,''
in \emph{ICML}, 2017.

\bibitem{pomerleau1989alvinn}
D.~A.~Pomerleau,
``ALVINN: An autonomous land vehicle in a neural network,''
in \emph{NeurIPS}, 1989.

\bibitem{ross2011dagger}
S.~Ross, G.~Gordon, and D.~Bagnell,
``A reduction of imitation learning and structured prediction to no-regret online learning,''
in \emph{AISTATS}, 2011.

\bibitem{florence2022ibc}
P.~Florence \emph{et al.},
``Implicit behavioral cloning,''
in \emph{CoRL}, 2022.

\bibitem{christiano2017preferences}
P.~F.~Christiano \emph{et al.},
``Deep reinforcement learning from human preferences,''
in \emph{NeurIPS}, 2017.

\bibitem{ajay2023compositional}
A.~Ajay \emph{et al.},
``Compositional foundation models for hierarchical planning,''
in \emph{NeurIPS}, 2023.

\bibitem{hu2022lora}
E.~J.~Hu \emph{et al.},
``LoRA: Low-rank adaptation of large language models,''
in \emph{ICLR}, 2022.

\bibitem{tobin2017domain}
J.~Tobin \emph{et al.},
``Domain randomization for transferring deep neural networks from simulation to the real world,''
in \emph{IROS}, 2017.

\bibitem{sermanet2018time}
P.~Sermanet \emph{et al.},
``Time-contrastive networks: Self-supervised learning from video,''
in \emph{ICRA}, 2018.

\bibitem{scholkopf2021causal}
B.~Sch\"olkopf \emph{et al.},
``Toward causal representation learning,''
\emph{Proc.\ IEEE}, 2021.

\bibitem{pearl2009causality}
J.~Pearl,
\emph{Causality: Models, Reasoning, and Inference},
2nd~ed.\ Cambridge University Press, 2009.

\bibitem{maniskill}
J.~Gu \emph{et al.},
``ManiSkill2: A unified benchmark for generalizable manipulation skills,''
in \emph{ICLR}, 2023.

\bibitem{khazatsky2024droid}
A.~Khazatsky \emph{et al.},
``DROID: A large-scale in-the-wild robot manipulation dataset,''
in \emph{RSS}, 2024.

\bibitem{fang2024rh20t}
H.-S.~Fang \emph{et al.},
``RH20T: A comprehensive robotic dataset for learning diverse skills in one-shot,''
in \emph{ICRA}, 2024.

\bibitem{o2024openx}
A.~Padalkar \emph{et al.},
``Open X-Embodiment: Robotic learning datasets and RT-X models,''
in \emph{ICRA}, 2024.

\bibitem{walke2023bridgedata}
H.~Walke \emph{et al.},
``BridgeData V2: A dataset for robot learning at scale,''
in \emph{CoRL}, 2023.

\end{thebibliography}
\end{document}